\pdfoutput=1

\documentclass[11pt]{article}

\usepackage{acl}

\usepackage{times}
\usepackage{latexsym}
\usepackage{listings}
\usepackage{tabularx}
\usepackage{multirow}
\usepackage{graphicx} 
\usepackage{kotex}
\usepackage{multicol}
\usepackage{enumitem} 
\usepackage[most]{tcolorbox} 
\usepackage{caption} 

\usepackage[T1]{fontenc}

\usepackage[utf8]{inputenc}

\usepackage{microtype}

\usepackage{inconsolata}

\usepackage[T1]{fontenc}

\usepackage[utf8]{inputenc}

\usepackage{microtype}

\usepackage{inconsolata}

\usepackage{amsmath} 

\title{Context-Aware LLM Translation System Using Conversation Summarization and Dialogue History}

\author{
Mingi Sung\textsuperscript{1,*}, 
Seungmin Lee\textsuperscript{2,*}, 
Jiwon Kim\textsuperscript{2}, 
Sejoon Kim\textsuperscript{1}\\
\textsuperscript{1}PwC Korea, Seoul, South Korea \\
\textsuperscript{2}Yonsei University, Seoul, South Korea \\
\textit{\{mingi.sung@pwc.com, elplaguister@yonsei.ac.kr, jwkim808@yonsei.ac.kr, sejoon.s.kim@pwc.com\}} \\
\small\textsuperscript{*}Equal contribution.
}

\begin{document}
\maketitle

\begin{abstract}
Translating conversational text, particularly in customer support contexts, presents unique challenges due to its informal and unstructured nature. We propose a context-aware LLM translation system that leverages conversation summarization and dialogue history to enhance translation quality for the English-Korean language pair. Our approach incorporates the two most recent dialogues as raw data and a summary of earlier conversations to manage context length effectively. We demonstrate that this method significantly improves translation accuracy, maintaining coherence and consistency across conversations. This system offers a practical solution for customer support translation tasks, addressing the complexities of conversational text. 
\end{abstract}

\section{Introduction}
The WMT 2024 Chat Shared Task addresses the unique challenges of translating conversational text, with a particular focus on customer support chats. Unlike formal or structured texts, conversations are typically spontaneous and casual, presenting several key challenges. First, the system must comprehend the dialogue's flow while accurately translating content from one language to another. Second, it is crucial to maintain logical continuity throughout entire conversations, preserving the context and intent of each exchange. Third, the task requires effectively handling the inherent noise and colloquial nature of chat data.

To tackle these challenges, we developed a context-aware LLM translation system that leverages both dialogue history and conversation summarization. Our approach is designed to maintain coherence and accuracy in translation by referencing two key elements: (1)‘History’ field: The two most recent dialogues of the target conversation, provided as raw data. (2)‘History Summary’ field: A concise summary (maximum 200 characters) of earlier conversations, excluding the two most recent dialogues.

We utilize both history and history summary in our approach for the following reasons. Dialogues often require multi-turn information for accurate understanding, as context within a single turn can be insufficient or misleading. Furthermore, while referencing all previous conversations would be ideal, it is often prohibited by the context length limitations of LLMs. Our method addresses these challenges by using recent parts of the conversation verbatim and summarizing earlier parts of the dialogue, effectively reducing context length while maintaining overall contextual information.

Our approach is informed by previous research demonstrating the effectiveness of context-aware models. Current study has shown that stimulating LLMs to memorize small dialogue contexts first and then recursively produce new memory using previous memory helps the chatbot generate highly consistent response \cite{wang2023recursively}. The History-Aware Hierarchical Transformer \cite{zhang-etal-2022-haht} also used historical information to improve the understanding of the current conversation context. The TiM (Think-in-Memory) framework \cite{liu2023think}, a LLM agent also recalls relevant thoughts from memory before generating response, and then integrates both historical and new thoughts to update the memory. By incorporating these insights, our system aims to produce translations that are not only accurate in language conversion but also maintain the coherent tone and appropriate word selection crucial in conversational contexts.

The Gemma-2-27B-it model \cite{team2024gemma} is used as the foundation for our translation system, specifically focusing on the English-Korean language pair. Our experiments demonstrated that incorporating recent dialogues and previous dialogue summaries significantly improved translation performance compared to methods that did not utilize this contextual information. We further refined our system by implementing more detailed instructions, which yielded additional improvements in translation accuracy. Furthermore, we used the GPT-4o mini \cite{gpt4o-mini} model for efficient conversation summarization. These combined methods resulted in substantial enhancements to overall translation quality, clearly demonstrating the effectiveness of our approach in boosting translation accuracy.

\section{Methodology}

\subsection{Data Preparation}

Among the five language pairs provided by the WMT 2024 Chat Shared Task, we selected the en-ko dataset for our experiments. The dataset provided by WMT consists of 16,122 training instances, 1,935 validation instances, and 1,982 test instances.

\newtcolorbox[auto counter, number within=section]{breakablebox}[1][]{colback=white, colframe=black, coltitle=black, boxrule=0.8mm, sharp corners, #1, breakable}

\begin{breakablebox}[label={sec:instructions}]

\noindent\textbf{Example 1} 
\vspace{0.5em}

\small 
\texttt{“source\_language”: “ko”, \\
\ \ \ “target\_language”: “en”, \\
\ \ \ “source”: “비밀번호 재설정 메일이 도착하지 않습니다.”, \\
\ \ \ “reference”: “I don't receive a password reset email.”, \\
\ \ \ “doc\_id”: “64619c16ab8523e90010b544”, \\
\ \ \ “client\_id”: “0015800001EMz0vAAD”, \\
\ \ \ “sender”: “customer”, \\
\ \ \ \textbf{“history”}: “As I understand you are unable to login to your account as it asks you to reset the password and you are not getting reset password email.”, \\
\ \ \ \ \ \ \ \ “제가 알기로는 비밀번호를 재설정하라는 메시지가 표시된 후 비밀번호 이메일을 재설정하지 않기 때문에 계정에 로그인할 수 없으십니다.” \\
\ \ \ \ \ \ \ \ “Am I correct?”, \\
\ \ \ \ \ \ \ \ “맞습니까?” \\
 \\
\ \ \ \textbf{“Instruction”}: “You are tasked with translating the following sentences from Korean to English. These sentences are part of conversations between a customer and a customer service agent.\textbackslash nWhen translating, keep the following instructions in mind:\textbackslash n- Provide only the translation of the ‘source’ text.\textbackslash n- Keep the translated text in a single line.\textbackslash n- The context involves a game user contacting a game company's customer service center online. Since the inquiries are typed, there may be many typos. Please translate with this in mind.\textbackslash n- Consider the summary of the previous conversation, referred to as ‘Dialogue Context’, if it is given.\textbackslash n- Refer to the context from the previous conversation if it is provided.\textbackslash n- Ensure your translations maintain the intended meaning and tone of the original dialogue.\textbackslash nDialogue Context: The customer, NAME-N, contacted PRS-ORG for help signing in and reported not receiving a password reset email.”, \\
\ \ \ \textbf{“History\_summary”}: “Dialogue Context: The customer, NAME-N, contacted PRS-ORG for help signing in and reported not receiving a password reset email.”, \\
\ \ \ \textbf{“System”}: “You are a professional translator fluent in both Korean and English.”
}

\normalsize 

\end{breakablebox}

During the data preparation process, we utilized several fields from the provided dataset and introduced two new ones to enhance context awareness. The original fields are \texttt{source\_language}, \texttt{target\_language}, \texttt{source}, \texttt{reference}, and \texttt{doc\_id}. The newly inserted fields are \texttt{history} and \texttt{history\_summary}. Example 1 shows the final preprocessed dataset used for model training. The \texttt{source\_language} and \texttt{target\_language} fields specify the language pair to be translated, \texttt{source} contains the text to be translated, \texttt{reference} provides the correct translation, and \texttt{doc\_id} is used to uniquely identify each conversation session. The \texttt{history} and \texttt{history\_summary} refer to the conversation context as described in the following sections.

\subsubsection{History}

The \texttt{history} field includes the raw data from the two previous dialogues of the targeted dialogue that needs to be translated. This information enables the model to capture the conversation's flow and maintain coherence in the generated translation.

\subsubsection{History Summary}

The \verb|history_summary| field contains a concise summary of earlier conversations, excluding the two most recent dialogues. It helps to understand the overall context and background of the current conversation. For summary generation, we used the GPT-4o mini model with a prompt that limits the summary to a maximum of 200 characters. This approach allows the model to focus on the essential part of the previous content without being overwhelmed by excessive details.

\subsubsection{System Prompt and Translation Instructions}

To ensure consistency in model training and translation tasks, we developed a prompt strategy. We defined a base prompt that positioned the model as a professional translator fluent in both Korean and English. We provided detailed guidelines for instruction, making the model focus on maintaining the context and tone of customer service dialogues, accounting for possible typos, and incorporating provided conversation summaries. To effectively guide the model, we supplied detailed prompts and instructions separately. The content of the system prompt and translation instructions can be found in the ‘system’ and ‘instruction’ fields in Example 1. This approach enhances translation quality by providing clear, context-specific guidance to the model.

\subsection{Context-Aware LLM Translation System Using Conversation
Summarization and Dialogue History}

After data preparation, we used the provided Chat Template for Gemma’s Instruct-Tuning to structure our data and fine-tuned the Gemma-2-27B-it model for translating customer support dialogues in Korean and English. We used DeepSpeed library to quantize the model and applied LoRA (Low-Rank Adaptation) for model compression. The key parameters of our training setup were as follows: per device train batch size of 4, gradient accumulation steps of 8, learning rate of 1.0e-4, 5 training epochs, cosine learning rate scheduler, warmup ratio of 0.1 and bfloat16 precision enabled. The GPU we used was NVIDIA H100 and the training process took about one hour to complete.

While optimizing our model's performance, we also addressed the unique challenges of dialogue translation. Context plays a pivotal role, significantly influencing the accuracy of interpreting and translating each turn. However, this importance presents a dual challenge. On one hand, preserving the conversation's history is crucial for coherent translations. On the other hand, as dialogues extend, managing this context becomes increasingly complex. Including all previous turns becomes impractical and can degrade the quality of subsequent translations.

To address these challenges, we implemented the following strategy:

\begin{itemize}
\item \textbf{Recent Dialogues (history)}: We utilized the history field to include the two most recent dialogue in their raw form. This preserves the immediate context necessary for accurate and coherent translations.

\item \textbf{Dialogue Context (history\_summary)}: For earlier parts, we provided a condensed summary of essential points, generated prior to inference. This helps the model grasp the broader context without being overwhelmed by excessive information \cite{bae-etal-2022-memory}.
\end{itemize}
This approach balances detailed immediate context with summarized background, allowing the model to capture both current dynamics and overall dialogue context.

Our prompt structure consists of three key components: a system role-play definition, a task instruction, and the sentence to be translated. This setup was critical for guiding the model's performance in customer support dialogue contexts. By structuring the context using natural language and leveraging the model's instruction-tuned capabilities, we aimed to enhance its ability to generate translations that are not only accurate but also contextually appropriate. This method allowed us to capture the natural flow and nuances of conversations more effectively.

\section{Experimental Results and Application}

\begin{table*}[h!]
\centering
\begin{tabular}{|c|c|c|c|}
\hline
\textbf{Team} & \textbf{Sentence (en$\rightarrow$ko)} & \textbf{Sentence (ko$\rightarrow$en)} & \textbf{Document} \\
\hline
\texttt{unbabel+it} & 93.39 & 96.31 & 93.21 \\
\texttt{DeepText\_Lab} & 91.35 & 95.71 & 90.04 \\
\texttt{DCUGenNLP} & 89.71 & 96.15 & 89.83 \\
\texttt{baseline} & 79.13 & 90.47 & 85.63 \\
\hline
\end{tabular}
\caption{Human Evaluation of Sentence and Document-Level Translation
(Test Dataset Results)}
\label{tab:team_comparison}
\end{table*}

\begin{table*}[h!]
\centering
\begin{tabular}{|c|c|c|c|c|}
\hline
\textbf{Team} & \textbf{COMET} & \textbf{chrF} & \textbf{BLEU} & \textbf{C-COMET-QE} \\
\hline
\texttt{unbabel+it} & 95.0 & 70.2 & 51.5 & 0.214 \\
\texttt{DeepText\_Lab} & 93.5 & 66.0 & 47.6 & 0.161 \\
\texttt{DCUGenNLP} & 92.3 & 59.8 & 39.4 & 0.158 \\
\texttt{baseline} & 87.6 & 48.9 & 26.0 & 0.041 \\
\hline
\end{tabular}
\caption{Automatic Evaluation Using Multiple Metrics
(Test Dataset Results)}
\label{tab:test_results}
\end{table*}

\begin{table*}[h!]
\centering
\small
\begin{tabular*}{\textwidth}{@{\extracolsep{\fill}}|l|c|c|c|c|c|}
\hline
\textbf{Configuration} & \textbf{Direction} & \textbf{COMET} & \textbf{chrF} & \textbf{BLEU} & \textbf{C-COMET-QE} \\
\hline
\multirow{2}{*}{\textbf{w/ recent dialogues and dialogue context}} & \texttt{en$\rightarrow$ko} & 0.916 & 51.52 & 32.97 & 0.138 \\
& \texttt{ko$\rightarrow$en} & 0.893 & 61.57 & 40.73 & - \\
\hline
\multirow{2}{*}{\textbf{w/o recent dialogues and dialogue context}} & \texttt{en$\rightarrow$ko} & 0.910 & 48.96 & 29.82 & 0.126 \\
& \texttt{ko$\rightarrow$en} & 0.889 & 59.65 & 38.86 & - \\
\hline
\multirow{2}{*}{\textbf{w/o prompt modification}} & \texttt{en$\rightarrow$ko} & 0.911 & 49.96 & 31.30 & 0.135 \\
& \texttt{ko$\rightarrow$en} & 0.894 & 61.15 & 40.64 & - \\
\hline
\end{tabular*}
\caption{Impact of Contextual Elements on Translation Performance
(Validation Dataset Results)}
\label{tab:validation_results_comparison}
\end{table*}

The performance of our translation model was evaluated through both human assessment and various automated metrics. Table 1 shows translation performance scores by human evaluation.  The ‘sentence' columns indicate the evaluation scores for translation quality at the individual sentence level, while the ‘Document' column reflects how well the translation maintains consistency and context across a full conversation. Our DeepTextLab team received notably high evaluations, with scores of 91.35 for translating sentences from English to Korean and 95.71 for translating sentences from Korean to English. Our team also received a score of 90.04 at the document level. These results demonstrate our system's strong performance at both sentence and document levels.

Besides human evaluation, the performance on the test dataset was also evaluated using several automated metrics, including COMET, chrF, BLEU, and Contextual-Comet-QE. The results are summarized in Table~\ref{tab:test_results}.
We achieved strong results across all metrics. The COMET score of 93.5 indicates high translation quality, while the chrF score of 66.0, BLEU score of 47.6, and Contextual-Comet-QE score of 0.161 demonstrate solid performance.

Beyond assessing overall performance, we explored how the inclusion of conversation history, history summaries, and detailed prompts influenced our model's translation quality. Table 3 illustrates the significant impact of these elements on various performance metrics, especially chrF and BLEU scores.  The absence of these contextual elements led to a significant decrease in translation quality, emphasizing the importance of context preservation and precise guidance in producing high-quality translations. 

In addition to the above evaluations, we also assessed our model's performance in terms of formality and lexical cohesion using the MuDA \cite{fernandes-etal-2023-translation} framework. The results of these assessments are presented in Table 4 and Table 5. For formality, the model achieved a precision score of 73.0, and a recall of 35.1, resulting in an overall F1-Score of 47.4. For lexical cohesion, the model demonstrated strong performance with a precision of 70.5, and a recall of 73.8, leading to an F1-Score of 72.1. 

\begin{table}[h!]
\centering
\begin{tabular}{|c|c|c|c|}
\hline
\textbf{Team} & \textbf{Precision} & \textbf{Recall} & \textbf{F1} \\
\hline
\texttt{unbabel+it} & 69.4 & 44.2 & 54.0 \\
\texttt{DeepText\_Lab} & 73.0 & 35.1 & 47.4 \\
\texttt{DCUGenNLP} & 25.5 & 18.2 & 21.2 \\
\texttt{baseline} & 50.0 & 10.4 & 17.2 \\
\hline
\end{tabular}
\caption{Formality Results}
\label{tab:formality_results}
\end{table}

\begin{table}[h!]
\centering
\begin{tabular}{|c|c|c|c|}
\hline
\textbf{Team} & \textbf{Precision} & \textbf{Recall} & \textbf{F1} \\
\hline
\texttt{unbabel+it} & 73.3 & 76.2 & 74.7 \\
\texttt{DeepText\_Lab} & 70.5 & 73.8 & 72.1 \\
\texttt{DCUGenNLP} & 73.5 & 68.3 & 70.8 \\
\texttt{baseline} & 66.1 & 65.1 & 65.6 \\
\hline
\end{tabular}
\caption{Lexical Cohesion Results}
\label{tab:lexical_cohesion_results}
\end{table}

\section{Conclusion}
We participated in the WMT English-Korean Chat Translation Task using the Gemma-2-27B-it model enhanced with dialogue history for context-aware translations. We effectively reduced the context length by summarizing earlier conversations and enhanced the model's translation performance by including the history of the two most recent dialogues and the summary of the previous dialogues, excluding the most recent two.

As a result, the translation performance has significantly improved, though there is still room for enhancement. Despite our team's high score, certain issues were identified in the generated translations. For instance, the Gemma 2 model occasionally produces translations in unexpected languages like Turkish, French, and Polish. This stems from the model's multilingual pretraining and presents an area for further exploration in future work.

\bibliography{acl_latex}

\end{document}